\documentclass{tlp-arxiv}

\usepackage{times}
\usepackage[english]{babel}
\usepackage{verbatim}
\usepackage{amssymb}
\usepackage{amsfonts}
\usepackage{amsmath}
\usepackage{graphicx}
\usepackage{latexsym}
\usepackage{url}
\usepackage{subfig}
\usepackage{hyperref}
\usepackage{quoting}
\quotingsetup{font=slshape, font+=small, indentfirst=false, leftmargin=15mm}
\usepackage{listings}
\usepackage{tikz}
\usepgflibrary{shapes.geometric}
\usetikzlibrary{shapes.geometric,shapes.misc,shapes}
\usetikzlibrary{arrows}
\usetikzlibrary{shadows}
\usetikzlibrary{fadings}
\usetikzlibrary{positioning}
\usetikzlibrary{calc}
\usetikzlibrary{fit}
\tikzset{vertex/.style={circle,minimum size=6mm,
      very thick,draw=black!50,
      top color=white,bottom color=black!20,inner sep=0pt}}

\def\fillCyl{%
	\fill[purple!15] let
		\p1 = ($(cyl.before top)!0.5!(cyl.after top)$),
		\p2 = (cyl.top),
		\p3 = (cyl.before top),
		\n1={veclen(\x3-\x1,\y3-\y1)},
		\n2={veclen(\x2-\x1,\y2-\y1)}
		in
		(\p1) ellipse (\n1 and \n2);
}

\newenvironment{dataflow}{%
\begin{tikzpicture}[
	every shadow/.style={opacity=.2},
	rectsContent/.style={
		draw,
		double copy shadow={left color=gray!20, right color=gray!50, middle color=white},
		left color=gray!10,
		right color=gray!30,
		middle color=white,
		outer sep=1ex, 
		xshift=-0.5ex,
	},
	rectsFittingBox/.style={
		inner sep=-0.5ex,
		xshift=0.5ex,
		yshift=0.5ex,
	},
	program/.style={
		draw,
		thick,
		rounded corners,
		drop shadow,
		left color=blue!10,
		right color=blue!25,
		middle color=white,
		inner sep=1ex,
	},
	storage/.style={
		draw,
		cylinder,
		alias=cyl,
		shape border rotate=90,
		aspect=0.15,
		outer sep=-0.5\pgflinewidth,
		left color=purple!10,
		right color=purple!30,
		middle color=white,
		drop shadow,
	},
	file/.style={
		draw,
		tape,
		tape bend top=none,
		tape bend height=1mm,
		left color=teal!10,
		right color=teal!30,
		middle color=white,
		drop shadow,
	},
	rect/.style={
		draw,
		rectangle,
		left color=gray!10,
		right color=gray!30,
		middle color=white,
	},
	font=\footnotesize,
	node distance=4mm,
	arrows=->,
	>=stealth,
]}{\end{tikzpicture}}

\newenvironment{flowchart}{%
\begin{tikzpicture}[
	rounded corners,
	every shadow/.style={opacity=.2},
	rect/.style={
		draw,
		rectangle,
		drop shadow,
		minimum height=5mm,
		top color=white,
		middle color=white,
		bottom color=red!80!blue!15,
	},
	diam/.style={
		draw,
		diamond,
		shape aspect=2,
		sharp corners,
		drop shadow,
		inner color=white,
		outer color=green!80!black!20,
	},
	font=\footnotesize,
	arrows=->,
	>=stealth,
]}{\end{tikzpicture}}

\newcommand{\NP}{\textbf{NP}}
\newcommand{\prob}[1]{\textsc{#1}}
\newcommand{\la}{\ensuremath{\longleftarrow}}
\newcommand{\dneg}{\ensuremath{\operatorname{not}}}

\title[D-FLAT: Declarative Problem Solving Using Tree Decompositions and ASP]%
{D-FLAT: Declarative Problem Solving Using Tree Decompositions and Answer-Set Programming}

\author[B.~Bliem, M.~Morak and S.~Woltran]
{
	BERNHARD BLIEM, MICHAEL MORAK and STEFAN WOLTRAN\\
	Institute of Information Systems 184/2\\
	Vienna University of Technology\\
	Favoritenstrasse 9--11, 1040 Vienna, Austria\\
	\email{[surname]@dbai.tuwien.ac.at}
}

\begin{document}

\maketitle

\begin{abstract}
In this work, we propose Answer-Set Programming (ASP) as a tool for
rapid prototyping of dynamic programming algorithms 
based on tree decompositions. In fact, many such algorithms 
have been designed, but only a few of them found their way
into implementation. The main obstacle is the lack of
easy-to-use systems which (i) take care of 
building a tree decomposition and (ii) provide an interface for declarative
specifications of dynamic programming algorithms. In this paper, we present
D-FLAT, a novel tool that relieves the user of having to handle all the
technical details concerned with parsing, tree decomposition, the handling
of data structures, etc. Instead, it is only the dynamic programming
algorithm itself which has to be specified in the ASP language. 
D-FLAT employs an ASP solver in order to compute the local solutions in the dynamic 
programming algorithm. In the paper, we give a few examples
illustrating the use of D-FLAT and describe the main features of the system.  
Moreover, we report experiments which show that ASP-based D-FLAT encodings for 
some problems outperform monolithic ASP encodings on instances of small 
treewidth.
\\
To appear in \emph{Theory and Practice of Logic Programming} (TPLP).
\end{abstract}

\section{Introduction}
\label{sec:introduction}

Computationally hard problems can be found in almost every area of computer 
science, hence the quest for general tools and methods that help designing 
solutions to such problems
is one of the central research challenges in our field.
One particularly successful 
approach 
is 
Answer-Set Programming~\cite{niem-99,mare-trus-99,BrewkaET11}%
---ASP, for short---%
where
highly sophisticated solvers \cite{dlv,GebserKKOSS11} provide a rich declarative 
language
to specify the given problem in an intuitive and succinct manner. 
On the other hand, the concept of dynamic programming \cite{Larson,Wagner}
denotes a general recursive
strategy where an optimal solution to a problem is defined in terms of optimal solutions to its
subproblems, thus
constructing a solution ``bottom-up'', from simpler to more complex problems.

One particular application of dynamic programming is the design of algorithms
which proceed along tree decompositions~\cite{DBLP:journals/jct/RobertsonS84}
of the problem at hand, or rather of its graph representation.
The significance of this approach is highlighted 
by Courcelle's seminal result (see, e.g., \cite{DBLP:books/el/leeuwen90/Courcelle90}) which 
states that every problem definable in monadic second-order logic
can be solved in linear time on structures of bounded treewidth;
formal definitions of these concepts will be provided in 
	Section~\ref{sub:treedecompositions}. 
This suggests a two-phased methodology for problem solving, where first
a tree decomposition of the given problem instance is obtained
and subsequently used in a second phase to solve the problem 
by a specifically designed  algorithm (usually employing dynamic programming) traversing the tree decomposition;
see, e.g., \cite{BodlaenderK08} for an overview of this approach.
Such tree decomposition based algorithms have
been successful in several applications including
constraint satisfaction problems such as \prob{Max-Sat}~\cite{Koster02},
and also bio-informatics~\cite{Xu05,CaiHLRS08}.

While good heuristics to obtain tree decompositions exist \cite{Dermaku08,Bodlaender10} and 
implementations thereof are available,
the actual implementation of dynamic programming algorithms that work on 
tree decompositions had, as yet, often to be done from scratch.
In this paper, we present a new 
method for declaratively
specifying the dynamic programming part, namely by means of 
ASP programs. As mentioned, dynamic programming relies on the evaluation of subproblems which 
themselves are often combinatorial in nature. Thanks to the Guess \& Check principle
of the ASP paradigm, using ASP to describe the treatment of subproblems is thus 
a natural choice and also separates our approach from existing systems (which 
we will discuss in the conclusion section of this paper). Using ASP as a 
language in order to describe the constituent elements of a dynamic programming 
algorithm obviously suggests to also employ sophisticated
off-the-shelf ASP systems 
as internal machinery for evaluating the specified algorithm. Thus, our approach 
not only takes full advantage of the rich syntax ASP offers to describe 
the dynamic programming algorithm, but also delegates
the burden of local computations to highly efficient ASP systems.

We have implemented our approach in the 
novel
system D-FLAT\footnote{%
(D)ynamic Programming (F)ramework with (L)ocal Execution of (A)SP on 
(T)ree Decompositions; available at
\url{http://www.dbai.tuwien.ac.at/research/project/dynasp/dflat/}.%
}
which takes care of the computation of tree decompositions 
and also 
handles the ``bottom-up'' data flow of the dynamic programming algorithm.\footnote{%
Concerning the decomposition, we employ
the htdecomp library of \cite{Dermaku08}
making our system amenable to hypertree decompositions, as well. 
This, in fact, allows to decompose arbitrary finite structures thus
extending the range of applicability for our system.}
All that is required of users is an idea for such an algorithm.
Hence, D-FLAT serves as a tool 
for rapid prototyping of dynamic programming algorithms but can also 
be considered for educational purposes. As well, ASP users are provided with
an easy-to-use interface to decompose problem instances---an issue which might
allow large instances of practical importance to be solved, which so far could 
not be handled by ASP systems.  To summarize, D-FLAT provides a new 
method for problem solving, combining the 
advantages
of 
ASP and tree decomposition based dynamic programming.
Some attempts to ease the specification of dynamic programming algorithms
already exist (see Section~\ref{sec:conclusion}), but we are not aware 
of any other system that employs ASP for dynamic programming on tree 
decompositions.

The structure of this paper is as follows: In Section~\ref{sec:preliminaries}, we first briefly 
recall logic programming under the answer-set semantics and 
provide
the necessary background for (hyper)tree decompositions and dynamic programming. In Section~\ref{sec:examples}, we introduce the features of D-FLAT step-by-step, providing ASP specifications of several 
standard (but also some novel) dynamic programming algorithms. 
Section~\ref{sec:systemdescription} gives some
system specifics, and Section~\ref{sec:experiments}
reports on a preliminary experimental evaluation.
In the conclusion of the paper, we address related 
work and give a brief summary and an outlook.

\section{Preliminaries}
\label{sec:preliminaries}

\subsection{Logic Programs and Answer-Set Semantics}
\label{sub:logicprogramsandanswersetsemantics}

The proposed system D-FLAT uses ASP to specify a dynamic programming algorithm.  
In this section, we will therefore briefly introduce the syntax and semantics of 
ASP.  The reader is referred to \cite{BrewkaET11} for a more thorough 
introduction.

A \emph{logic program} $\Pi$ consists of a set of \emph{rules} of the form
\[a_1 \lor \dots \lor a_k \la b_1, \dots, b_m, \dneg b_{m+1}, \dots, \dneg 
b_n.\]
We call $a_1, \dots, a_k$ and $b_1, \dots, b_n$ \emph{atoms}. A \emph{literal} is either an 
atom or its default negation which is obtained by putting $\dneg$ in front.
For a rule $r \in \Pi$, we call $h(r) = \{a_1, \dots, a_k\}$ its \emph{head} and $b(r) 
= \{b_1, \dots, b_n\}$ its \emph{body} which is further divided into a \emph{positive body}, 
$b^+(r) = \{b_1, \dots, b_m\}$, and a \emph{negative body}, $b^-(r) = \{b_{m+1}, \dots, 
b_n\}$.
Note that the head may be empty, in which case we call $r$ an \emph{integrity 
constraint}.
If the body is empty, $r$ is called a \emph{fact}, and the $\la$ symbol can be 
omitted.

A rule $r$ is satisfied by a set of atoms 
$I$ (called an \emph{interpretation})
iff $I \cap 
h(r) \neq \emptyset$ or $b^-(r) \cap I \neq \emptyset$ or $b^+(r) \setminus I 
\neq \emptyset$.
$I$ is a \emph{model} of a set of rules iff it satisfies each rule.
$I$ is an \emph{answer set} of a program $\Pi$ iff it is a subset-minimal model of
the program $\Pi^I =\{h(r) \la b^+(r) \mid r \in \Pi, b^-(r) \cap I =
\emptyset\}$, called the 
\emph{Gelfond-Lifschitz reduct} \cite{gelf-lifs-91} of $\Pi$ w.r.t.\ $I$.

In this paper, we use the input language of Gringo 
\cite{clasp-guide}
for logic
programs. Atoms in this language are predicates whose arguments can be either
variables or ground terms. Such programs can be seen as abbreviations of
variable-free programs, where all the variables are instantiated (i.e., replaced
by ground terms). The process of instantiation is also called \emph{grounding}.
It results in a propositional program which can be represented as a set of rules 
which adhere to the above definition.

As an example, an instance for the \prob{3-Col} problem\footnote{%
\prob{3-Col} is defined as the problem of deciding whether, for a given graph 
$(V,E)$, there exists a 3-coloring, i.e., a mapping $f: V \to 
\{\text{red},\text{green},\text{blue}\}$ s.t.\ for each edge $(a,b) \in E$ it 
holds that $f(a) \neq f(b)$.}
(covered in Section~\ref{sec:graph-coloring}) consists of a set of vertices and 
a set of edges, so the following set of facts represents a valid instance:
\begin{lstlisting}[numbers=none]
vertex(a). vertex(b). vertex(c). vertex(d). vertex(e).
edge(a,b). edge(a,c). edge(b,c). edge(b,d). edge(c,d). edge(d,e).
\end{lstlisting}
The following logic program solves \prob{3-Col} for instances of this form:
\begin{lstlisting}
color(red;green;blue).
1 { map(X,C) : color(C) } 1 :- vertex(X).
:- edge(X,Y), map(X,C), map(Y,C).
\end{lstlisting}
Solving \prob{3-Col} for an instance amounts to grounding this encoding together 
with the instance as input and then solving the resulting ground program.
Line~1 of the encoding is expanded by the grounder to three facts which state 
that ``red'', ``green'' and ``blue'' are colors.  Line~2 contains a cardinality 
constraint in the head and is conceptually expanded to
\begin{lstlisting}[numbers=none]
1 { map(X,red), map(X,green), map(X,blue) } 1 :- vertex(X).
\end{lstlisting}
before the grounder expands this rule by substituting ground terms for 
\verb!X!.
Roughly speaking, a cardinality constraint $l \{ L_1, \dots, L_n \} u$ is 
satisfied by an interpretation $I$ iff at least $l$ and at most $u$ of the literals $L_1, \dots, L_n$ are 
true in $I$. Therefore, the rule in question expresses a choice of exactly one of 
\verb!map(X,red)!, \verb!map(X,green)!, \verb!map(X,blue)!, for any vertex 
\verb!X!. Finally, the integrity constraint in line~3 of the encoding ensures 
that no answer set maps the same color to adjacent vertices.
For space reasons, we refer the reader to \cite{clasp-guide} for more details on 
the input language.

\vspace{-2mm}

\subsection{Hypertree Decompositions}
\label{sub:treedecompositions}

Tree decompositions and treewidth, originally defined in
\cite{DBLP:journals/jct/RobertsonS84}, are a well known tool to tackle
computationally hard problems (see, e.g.,
\cite{DBLP:journals/actaC/Bodlaender93,DBLP:conf/sofsem/Bodlaender05} for an
overview). Informally, treewidth is a measure of the cyclicity of a graph and many
\NP-hard problems become tractable if the treewidth is bounded.
The intuition behind tree decompositions is obtaining a tree from a (potentially 
cyclic) graph by subsuming multiple vertices under one node and thereby 
isolating the parts responsible for the cyclicity.

Several problems are better represented as hypergraphs for which the concept of 
tree decomposition can be generalized,
see, e.g., \cite{GottlobLS02}. In the following, we therefore define hypertree 
decompositions, of which tree decompositions are a special case.

  A \emph{hypergraph} is a pair $H=(V,E)$ with a set $V$ of vertices and a set
  $E$ of hyperedges. A hyperedge $e \in E$ is itself a set of vertices with $e
  \subseteq V$.
  A \emph{hypertree decomposition} of a hypergraph $H=(V,E)$ is a pair
  $\mathit{HD}=\langle T,\chi\rangle$, where $T=(N,F)$ is a (rooted) tree,
  with $N$ being its set of nodes and $F$ its set of edges,
  and
  $\chi$ 
  is a labeling function such that for each node $n \in
  N$ the so-called \emph{bags} $\chi(n)\subseteq V$ of $\mathit{HD}$ 
	meet the following requirements:

\vspace{-1mm}
  \begin{enumerate}
  \item for every 
		$v\in V$ there exists a node $n\in N$ such that
    $v\in\chi(n)$,
  \item for every 
		$e\in E$ there exists a node $n\in N$ such that
    $e\subseteq\chi(n)$,
  \item for every 
	$v\in V$ the set $\lbrace n\in N\mid
    	v\in\chi(n)\rbrace$ induces a connected subtree of $T$.
  \end{enumerate}
\vspace{-1mm}
  
\begin{figure}
\caption{A graph and a corresponding semi-normalized tree decomposition}
\label{fig:td-example}
\subfloat{\begin{tikzpicture}[scale=0.5]
\tikzstyle{every node}=[draw,circle,fill=green!10!blue!5];
\tikzstyle{every path}=[draw,thick];
\node (a) at (-0.5,0) {a};
\node (b) at (1,1) {b};
\node (c) at (1,-1) {c};
\node (d) at (2.5,0) {d};
\node (e) [right of=d] {e};
\draw (a) to (b);
\draw (a) to (c);
\draw (b) to (c);
\draw (b) to (d);
\draw (c) to (d);
\draw (d) to (e);
\end{tikzpicture}}
\qquad
\subfloat{\begin{tikzpicture}[scale=0.5]
\tikzstyle{every node}=[draw,rounded corners,fill=green!10!blue!5];
\tikzstyle{every path}=[draw,thick];
\node at (-1.5,0) (leaf1) {$\{a,b,c\}$};
\node at (1.5,0) (leaf2) {$\{d,e\}$};
\node at (-1.5,1.5) (exchange1) {$\{b,c,d\}$};
\node at (1.5,1.5) (exchange2) {$\{b,c,d\}$};
\node at (0,3) (join) {$\{b,c,d\}$};

\draw (join) to (exchange1);
\draw (join) to (exchange2);
\draw (exchange1) to (leaf1);
\draw (exchange2) to (leaf2);
\end{tikzpicture}}
\end{figure}
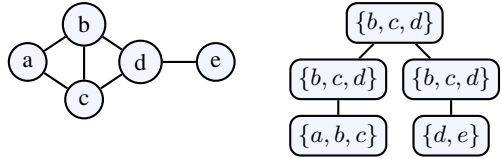

A hypertree decomposition $\langle T,\chi\rangle$ is called \emph{semi-normalized} if each node 
$n$ in $T$ has at most two children---nodes with zero (resp.\ one, two) children are
called leaf (resp.\ exchange, join) nodes---and for a join node $n$ with children $n_1$ and $n_2$,
$\chi(n)=\chi(n_1)=\chi(n_2)$ holds.%
\footnote{A few terminological remarks: The reason we use the term 
``\emph{semi}-normalized'' is that the (more restrictive) concept of 
\emph{normalized} (also called \emph{nice}) tree decompositions appears in the 
literature. Normalized tree decompositions are also semi-normalized. Moreover, 
hypertree decompositions generalize the notion of tree decompositions to the 
case of hypergraphs. Therefore, we often use both terms interchangeably if we 
are dealing with ordinary graphs, even though, strictly speaking, D-FLAT always 
produces a hypertree decomposition.}
Figure~\ref{fig:td-example} depicts a semi-normalized tree decomposition for the 
example graph given by the problem instance in 
Section~\ref{sub:logicprogramsandanswersetsemantics}.

	For defining the \emph{width} of a hypertree decomposition $\langle(N,F),\chi\rangle$ of 
	a hypergraph $(V,E)$, we need an additional labeling function $\lambda:N\rightarrow 2^E$
	such that, for every $n\in N$, 
	$\chi(n)\subseteq\bigcup_{e\in\lambda(n)}e$.
  A hypertree decomposition is called \emph{complete} when for each hyperedge
  $e\in E$ there exists a node $n\in N$ such that $e\in\lambda(n)$.
  The \emph{width} of a hypertree decomposition is then the maximum $\lambda$-set
  size over all its nodes. The minimum width over all possible hypertree
  decompositions is called the \emph{generalized hypertree width}.
	The idea of this parameter is to capture the inherent difficulty of the 
	given problem, such that the smaller the generalized hypertree width
	(i.e., the level of cyclicity), the easier
	the problem is to solve, see, e.g., \cite{GottlobLS02}.
  If $G=(V,E)$ is an ordinary graph, then the width of a tree decomposition $\langle T,\chi\rangle$
  is defined differently, namely as the maximum bag size $\lvert \chi(n) \rvert$, minus one.
%
For a given hypergraph, it is \NP-hard to compute a hypertree
decomposition of minimum width.  
However, efficient heuristics have been
developed that offer good approximations (cf.  \cite{Dermaku08,Bodlaender10}).
It should be 
noted that in general there exist many possible 
hypertree decompositions for a given hypergraph, and there always exists at 
least one: 
the degenerated hypertree decomposition 
consisting of just a single node which covers the entire hypergraph. D-FLAT 
expects the dynamic programming algorithms provided by the user to work on any 
semi-normalized hypertree decomposition, so users do not have to worry 
about (nor can they currently determine) 
which decomposition is 
generated. This is up to the heuristic methods 
of the htdecomp library, which attempt to construct decompositions of small width.

\subsection{Dynamic Programming on Hypertree Decompositions}
\label{sub:dynamicprogramming}

The value of hypertree decompositions is that their size is only linear in the 
size of the given graph. Moreover, they provide a suitable structure to design 
dynamic programming algorithms for a wide range of problems. These algorithms 
generally start at the leaf nodes and traverse the tree to the root, whereby at each node 
a set of partial solutions is generated by taking those solutions into account 
that have been computed for the child nodes.  The most difficult part in 
constructing such an algorithm is to identify an appropriate data structure to 
represent the partial solutions at each node: On the one hand, this data 
structure must contain sufficient information so as to
compute the representation of the partial solutions at each node from the
corresponding representation at the child node(s). On the other hand, the size
of the data structure should only depend on the size of the bag (and not on the
total size of the  instance to solve). 
%
\begin{figure}
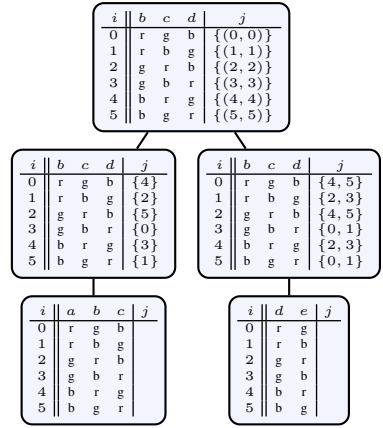

\caption{The \prob{3-Col} tuple tables for the instance and tree decomposition 
given in Figure~\ref{fig:td-example}}
\label{fig:3col-example}
\begin{tikzpicture}[scale=1.3]
\tikzstyle{every node}=[draw,rounded corners,font=\tiny,fill=green!10!blue!5];
\tikzstyle{every path}=[draw,thick];
\node at (-1,0) (leaf1) {\input{leaf1}};
\node at (1,0) (leaf2) {\input{leaf2}};
\node at (-1,1.5) (exchange1) {\input{exchange1}};
\node at (1,1.5) (exchange2) {\input{exchange2}};
\node at (0,3) (join) {\input{join}};

\draw (join) to (exchange1);
\draw (join) to (exchange2);
\draw (exchange1) to (leaf1);
\draw (exchange2) to (leaf2);
\end{tikzpicture}
\end{figure}
For this purpose, at each tree decomposition node we maintain a data structure 
which we call \emph{table}. A table contains rows which we call \emph{tuples} 
(i.e., mappings that assign some value to each current bag element), in which we 
store the partial solutions. Additionally, each tuple in non-leaf nodes may be 
associated with tuples from child nodes by means of pointers.  These pointers 
determine which child tuples a tuple has originated from and are used to 
construct complete solutions out of the respective partial solutions when the 
computation of all tables is finished.
\footnote{If only the decision problem needs to be solved, these pointers are
not necessary; in general, the algorithm and the structure of the tables depend
on the problem type (cf.\ \cite{GrecoS10,GottlobGS09}).}

We illustrate this idea for the \prob{3-Col} problem.  Consider a given graph 
$G$ and a corresponding tree decomposition. For each node $n$ in the decomposition, we 
basically compute the valid colorings of the subgraph of $G$ induced  by 
$\chi(n)$, i.e., by the current bag, and store these colorings in the rows of 
the table associated with $n$.  Hence each tuple of node $n$'s table assigns 
either ``r'', ``g'' or ``b'' to each vertex that is contained in the current bag 
$\chi(n)$. So far, the steps taken by the dynamic programming algorithm in $n$ 
do not differ from a general Guess \& Check approach to \prob{3-Col}.
However, when computing the colorings for an exchange node $n$, we start from
the colorings for the child node $n'$ of $n$ and adapt them with respect to
the new vertices $\chi(n)\setminus\chi(n')$.
In addition, we also keep track of which tuples of $n'$ give rise to 
which newly calculated tuples of $n$.  Figure~\ref{fig:3col-example} depicts the respective 
tables of each node of the semi-normalized tree decomposition of 
Figure~\ref{fig:td-example}.  In each table, column $i$ contains an index 
for each tuple, that can be used in column $j$ of a potential parent node to 
reference it. For exchange nodes, an entry in column $j$
is a set of pointers to such child tuples. For join nodes, an entry in column $j$ is a
set of pairs $(l,r)$, where $l$ (resp.\ $r$) references a tuple of the left
(resp.\ right) child node.  Using these pointers, it is possible to enumerate
all complete solutions to the entire problem by a final top-down traversal.

Note that the size of the tables only depends on the width of the tree
decomposition, and the number of tables is linear in the size of the input
graph. Thus, when the width is bounded by a constant, the search space for each
subproblem remains constant as well, and the number of subproblems only grows by
a linear factor for larger instances.

\section{D-FLAT by Example}
\label{sec:examples}

In this section, we introduce the usage of the D-FLAT system by means of 
specific example problems. We begin with a relatively simple case to illustrate 
the basic functionality of the system and then continue with more complex 
applications.
In the course of this, we will gradually introduce the features of D-FLAT and 
the special predicates responsible for the communication between D-FLAT and the 
user's programs. These predicates are summarized in 
Table~\ref{tab:special_predicates} for future reference.
A general description of the 
system and its possible applications 
is delegated to
Section~\ref{sec:systemdescription}.

\begin{table}
\caption{Reserved predicates for the communication of the user's programs with 
D-FLAT}
\label{tab:special_predicates}
\begin{tabular}{p{4cm}p{8cm}}
\hline
\hline
Input predicate & Meaning\\
\hline

\verb!current(V)! & $V$ is an element of the current bag.\\
\noalign{\vspace{1ex}}

\verb!introduced(V)! & $V$ was introduced in the current bag.\\
\noalign{\vspace{1ex}}

\verb!removed(V)! & $V$ was removed in the current bag.\\
\noalign{\vspace{1ex}}

\verb!childTuple(I)! & $I$ is the identifier of a child tuple. (Only in exchange 
nodes)\\
\noalign{\vspace{1ex}}

\verb!childTupleL(I)!, \verb!childTupleR(I)! & $I$ is a tuple from the left 
resp.\ right child node.\newline (Only in join nodes)\\
\noalign{\vspace{1ex}}

\verb!mapped(I,X,Y)! & Child tuple $I$ assigns to $X$ the value $Y$.\\
\noalign{\vspace{1ex}}

\verb!childCost(I,C)! & $C$ is the total cost of the partial solution 
corresponding to the child tuple $I$.\\
\noalign{\vspace{1ex}}

\verb!root! & Indicates that the current node is the root node.\newline
(Only in exchange nodes)\\

\hline
Output predicate & Meaning\\
\hline

\verb!map(X,Y)! & Assign the value $Y$ to the current bag element $X$.\\
\noalign{\vspace{1ex}}

\verb!chosenChildTuple(I)! & Declare $I$ to be the identifier of the child tuple 
that is the predecessor of the currently described one.\newline (Only in 
exchange nodes; not necessary for decision problems)\\
\noalign{\vspace{1ex}}

\verb!chosenChildTupleL(L)!, \verb!chosenChildTupleR(R)! & Declare $L, R$ to be 
the identifiers of the preceding child tuples from the left resp.\ right child 
node.\newline (Only in join nodes; not necessary for decision problems)\\
\noalign{\vspace{1ex}}

\verb!cost(C)! & Let $C$ be the total cost of the current partial 
solution.\newline (Only required when solving an optimization problem)\\
\noalign{\vspace{1ex}}

\verb!currentCost(C)! & Let $C$ be the local cost of the current tuple.\newline 
(Only required in exchange nodes when solving an optimization problem and using 
the default join implementation)\\
\hline
\hline
\end{tabular}
\end{table}

\vspace{-3mm}
\subsection{Graph Coloring}
\label{sec:graph-coloring}
As a first example of how to solve an \NP-complete problem using D-FLAT, we 
consider \prob{3-Col}, the 3-colorability problem for graphs.
We have already given an encoding for \prob{3-Col} in Section~\ref{sub:logicprogramsandanswersetsemantics}.
Since this program, together with the instance as input, solves the whole 
problem, we call it a \emph{monolithic encoding}. 
As \prob{3-Col} is 
fixed-parameter tractable w.r.t.\ the treewidth, we can take advantage of low 
treewidths by solving the problem with a dynamic programming algorithm that 
operates on a tree decomposition of the original input graph.
We have sketched the dynamic programming algorithm for \prob{3-Col} in 
Section~\ref{sub:dynamicprogramming}.

D-FLAT now provides a means to specify this dynamic programming algorithm in the ASP 
language. It reads the instance, stores its graph representation, and from this 
it constructs a semi-normalized tree decomposition.
In order for D-FLAT to obtain a graph representation from the instance, the user 
only needs to specify on the command line which predicates indicate (hyper)edges 
in the graph representation (in this case only \verb!edge/2!).

Following the idea of dynamic programming, we wish to compute a table
for each
tree decomposition node, such that a table can be computed using the 
tables of the child nodes; i.e., we perform the calculations in a bottom-up 
manner.
Since semi-normalized tree decompositions only allow for two kinds of nodes---%
\emph{exchange nodes}, which have one child, and \emph{join nodes}, 
which have exactly two children with the same contents as the join node---it 
suffices to provide D-FLAT with an ASP program that computes the table for 
exchange nodes (\emph{exchange program} for short) and with a program that 
computes the table for join nodes (\emph{join program} for short).%
\footnote{In fact,
D-FLAT produces tree decompositions where all leaves as well as the root node 
have an empty bag.  Empty leaf nodes have the advantage that they do not involve 
any problem-specific
computation (at least in the use cases we considered so far) but just deliver
the empty tuple (with cost $0$) to its parent node. (This default behavior is 
mirrored in D-FLAT in the
sense that it currently only supports user-specified ASP programs for exchange 
and join nodes).
On the other hand, having an empty root node guarantees that
final actions of
the dynamic programming algorithm 
can always be specified in the program for the exchange nodes.}
These two programs are all that is required of the user to solve a problem.
D-FLAT traverses the tree decomposition in post-order and, for each node, 
invokes the ASP solver with the proper program, i.e., with the exchange program 
for exchange nodes and with the join program for join nodes.
\begin{figure}
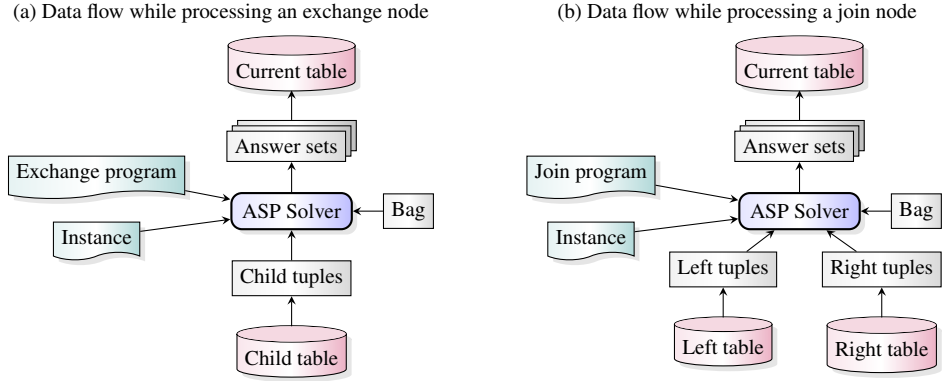

\caption{Workflow for calculating a tree decomposition node's table}
\label{fig:dataflow}
\subfloat[Data flow while processing an exchange node]{
		\input{exchange-dataflow}}
\hfill
\subfloat[Data flow while processing a join node]{
		\input{join-dataflow}
}
\end{figure}
An illustration of 
such single
steps is given in 
Figure~\ref{fig:dataflow}.
The input for the exchange or join program 
consists of
\begin{itemize}
\item the original problem instance as supplied by the user,
\item the tuples from the child nodes as a set of facts,
\item the current bag, i.e., the current, introduced and 
removed vertices, as a set of facts.
\end{itemize}
The answer sets then constitute the current node's table. D-FLAT takes care of 
processing them and filling the appropriate data structures. Once all new tuples 
have thus been stored, it proceeds to the next node. This procedure continues 
until the root has been processed.

We now present an exchange program for \prob{3-Col}.
Regarding the input that is 
supplied by D-FLAT, the set of current, introduced and removed vertices is 
given by 
predicates \verb!current/1!, \verb!introduced/1! and 
\verb!removed/1!, respectively.  Each child tuple has some identifier $I$ and is 
declared by the fact \verb!childTuple(I)!.  The corresponding mapping is given 
by facts of the form \verb!mapped(I,X,C)!, which signifies that in the tuple $I$ 
the vertex $X$ is assigned the color $C$. (The predicate name ``mapped'' was 
chosen in analogy to the present tense term ``map''.)
\begin{lstlisting}
color(red;green;blue).
1 { chosenChildTuple(I) : childTuple(I) } 1.
map(X,C) :- chosenChildTuple(I), mapped(I,X,C), current(X).
1 { map(X,C) : color(C) } 1 :- introduced(X).
:- edge(X,Y), map(X,C), map(Y,C).
\end{lstlisting}
The program above is based on the intuition that in an exchange node each child tuple 
gives rise to a set of new tuples, such that the new tuples coincide on the 
coloring of common vertices~(line~3), and the coloring of introduced vertices is 
guessed~(line~4), followed by a check~(line~5) which uses the \verb!edge/2!  
predicate of the problem instance.

Each answer set here constitutes exactly one new tuple in the current node's 
table.  The output predicate \verb!map/2! is used to specify the partial 
coloring of the new tuple, just like it was used in the monolithic encoding. It 
must assign a color to each vertex in the current node.  Another output 
predicate recognized by D-FLAT is \verb!chosenChildTuple/1!. It indicates which 
child tuple a partial coloring (characterized by an answer set) corresponds to 
and must, of course, only have one child tuple identifier as its extension. This 
predicate is required for the reconstruction of complete solutions after all 
tables have been computed and can therefore be omitted if just the decision 
problem should be solved.

It still remains to provide D-FLAT with a program for processing the other kind 
of nodes, viz., the join nodes. A join program for this purpose could look like 
this:
\begin{lstlisting}
1 { chosenChildTupleL(I) : childTupleL(I) } 1.
1 { chosenChildTupleR(I) : childTupleR(I) } 1.
:- chosenChildTupleL(L), chosenChildTupleR(R),                       mapped(L,X,A), mapped(R,X,B), A != B.        % Only join equal tuples
map(X,V) :- chosenChildTupleL(L), mapped(L,X,V).
\end{lstlisting}
Here an answer set obviously must state \emph{for each} of the two child nodes
the chosen preceding tuple (indicated by \verb!chosenChildTupleL! and 
\verb!chosenChildTupleR!).
In the case of \prob{3-Col}, two tuples match (i.e., are joined to a new tuple) 
iff their partial assignments coincide (line~3). The tuple resulting from such a 
join is then also equal to each of the two matching child tuples (line 4).

This particular join behavior is so basic that it also applies to other common 
problems.  Therefore, D-FLAT offers a (quite efficient) default implementation 
of this behavior, which users can resort to (if it suffices) instead of 
writing their own join programs.


We would like to 
point out here the difference between the 
role of exchange and join nodes. 
In 
particular, one might wonder why for join nodes there is a default 
implementation that can often be used, whereas there is none for exchange nodes.
This difference is due to the idea of dynamic programming which, on the one 
hand, involves solving partial problems and, on the other hand, requires 
combining partial solutions (which is clearly separated, thanks to the
concept of semi-normalized tree decompositions). Obviously, the combining is explicitly done in join 
nodes, whereas usually the solving of partial problems is performed in exchange 
nodes, which therefore require problem-specific code. Exchange nodes 
\emph{always} require problem-specific code since the entire problem can be seen 
as a single degenerated exchange node.
Hence, parts of exchange programs resemble monolithic encodings very much, 
even though these are technically irrelevant to D-FLAT.
Indeed, exchange programs can be seen as more general than monolithic encodings 
since a correct exchange program must in principle also solve the entire 
problem.
In some cases however, knowledge of the problem is also needed in join nodes, as the 
next subsection will show.

\subsection{Boolean Satisfiability}
\label{sec:sat}

Instances for the 
satisfiability problem (\prob{Sat}) are given by the 
predicates \verb!pos(C,V)! and \verb!neg(C,V)!, denoting that the propositional 
variable $V$ occurs positively resp.\ negatively in the clause $C$. Each clause
and variable is declared using the predicates
\verb!clause/1! and \verb!atom/1!, respectively.
%
The following monolithic encoding solves such instances:
\begin{lstlisting}
1 { map(A,true), map(A,false) } 1 :- atom(A).
sat(C) :- pos(C,A), map(A,true).
sat(C) :- neg(C,A), map(A,false).
:- clause(C), not sat(C).
\end{lstlisting}
We can obtain a graph representation of a \prob{Sat} instance by constructing 
its incidence graph, i.e, the graph obtained by considering the clauses and variables as vertices 
and connecting a clause vertex with a variable vertex by an edge iff the 
respective variable occurs in the respective clause. Again, in order for D-FLAT 
to transform the input into a graph representation, the user is only required to 
state that \verb!pos/2!  and \verb!neg/2! indicate edges.

A dynamic programming algorithm for the model counting problem working on tree 
decompositions of the incidence graph is given in \cite{Samers10}.
We now present a possible ASP encoding for the exchange node that follows that 
work's general idea of such an algorithm and
generalizes it for semi-normalized tree decompositions.
It 
should be noted that we primarily assign truth values to current propositional 
atoms like in the monolithic encoding, but as a consequence of this we also 
assign true or false to each current clause depending on whether or not it is 
satisfied by the partial interpretation represented by the tuple. We need this 
information on the status of a clause because, when a clause is removed, all 
tuples not satisfying this clause must be eliminated.
\begin{lstlisting}
1 { chosenChildTuple(I) : childTuple(I) } 1.
:- clause(C), removed(C), chosenChildTuple(I), not mapped(I,C,true).
map(X,true) :- chosenChildTuple(I), mapped(I,X,true), current(X).
1 { map(A,true), map(A,false) } 1 :- atom(A), introduced(A).
map(C,true) :- pos(C,A), current(C), map(A,true).
map(C,true) :- neg(C,A), current(C), map(A,false).
map(X,false) :- current(X), not map(X,true).
\end{lstlisting}
The mentioned elimination of non-satisfying truth assignments is performed by 
the check in line~2. The child tuple's partial interpretation is retained 
(lines~3 and 7) and extended by a guess on introduced atoms (line~4).  If a 
clause was satisfied at some point before, it remains so due to line 3, whereas 
lines~5 and 6 mark clauses as satisfied by the current partial assignment.

For the \prob{Sat} problem we cannot use D-FLAT's default implementation for 
join nodes. The reason is that, in order to be joined, two tuples do not need to 
coincide on the values they assign to clauses, but only on the assignments for 
atoms. If, for a given truth assignment on the current atoms, a clause is 
satisfied by either child tuple, it is also satisfied by the tuple resulting 
from joining the two. The following join program follows this idea:
\begin{lstlisting}
1 { chosenChildTupleL(I) : childTupleL(I) } 1.
1 { chosenChildTupleR(I) : childTupleR(I) } 1.
:- mapped(L,A,true), mapped(R,A,false), atom(A), chosenChildTupleL(L), chosenChildTupleR(R).
:- mapped(L,A,false), mapped(R,A,true), atom(A), chosenChildTupleL(L), chosenChildTupleR(R).
map(X,true) :- chosenChildTupleL(I), mapped(I,X,true).
map(X,true) :- chosenChildTupleR(I), mapped(I,X,true).
map(X,false) :- current(X), not map(X,true).
\end{lstlisting}

\subsection{Minimum Vertex Cover}

Encodings as outlined above suffice to instruct D-FLAT to solve the respective 
enumeration problems (``What are the solutions?''), counting problems (``How 
many solutions exist?'') and decision problems (``Is there a solution?'').  
Often it is also desired to solve optimization problems, for which additional 
information regarding the cost of a (partial) solution must be supplied in the 
encodings. As an example, we briefly introduce how the ``drosophila'' of 
fixed-parameter algorithms, the \prob{Minimum Vertex Cover} problem, can be 
solved. Instances are again given by \verb!vertex/1! and \verb!edge/2!.  As 
before, we begin with a monolithic encoding for comparison:
\begin{lstlisting}
1 { map(X,in), map(X,out) } 1 :- vertex(X).
:- edge(X,Y), map(X,out), map(Y,out).
cost(C) :- C = #count{ map(X,in) }.
#minimize[ cost(C) = C ].
\end{lstlisting}
By means of the \verb!#minimize! statement, we leave it to the ASP solver to 
filter out suboptimal solutions. However, it is a mistake to use such an 
optimization statement when writing the exchange program for D-FLAT because 
then one would filter out tuples whose local cost might exceed that of others 
but which in the end would yield a better global solution.

An exchange program for \prob{Minimum Vertex Cover} could look like this:
\begin{lstlisting}
1 { chosenChildTuple(I) : childTuple(I) } 1.
map(X,Y) :- current(X), chosenChildTuple(I), mapped(I,X,Y).
1 { in(X), out(X) } 1 :- introduced(X).
map(X,in) :- in(X).
map(X,out) :- out(X).
:- edge(X,Y), map(X,out), map(Y,out).
currentCost(C) :- C = #count{ map(_,in) }.
cost(C) :- chosenChildTuple(I), childCost(I,CC), IC = #count{ in(_) },    C = CC + IC.
\end{lstlisting}
The \verb!cost/1! predicate is recognized by D-FLAT and specifies the cost of 
the partial solution (obtained by extending the current tuple with its 
predecessors, recursively). This number is computed in line~8 by adding to the 
preceding tuple's cost (which is provided by D-FLAT as \verb!childCost/2!) the 
cost which is due to introduced vertices---in this case, the number of vertices 
guessed as \verb!in!.

A peculiarity when using the default join implementation (as in this example) is 
that in each tuple we not only need to store the cost of the corresponding 
partial solution but also the cost of the tuple itself, declared by 
\verb!currentCost/1!, because upon joining two coinciding tuples and adding the 
total costs of their associated partial solutions, the portion of the cost that 
is due to the coinciding child tuples is counted twice and must thus be 
subtracted from the sum. This is why D-FLAT also recognizes the 
\verb!currentCost/1! predicate. When implementing the join program manually, 
this predicate is useless, as calculating the cost of a tuple resulting from a 
join is then up to the user.

Note that, as soon as the root of the tree decomposition has been processed by 
D-FLAT, it possesses all the information required for determining the cost and 
number of optimal solutions in constant time (i.e., without an additional tree 
traversal), as well as for enumerating all optimal solutions with linear delay 
by traversing the tree top-down, following each tuple's pointers to its 
predecessors to construct solutions.  The processing of an optimization problem 
instance is no different from that of problems without optimization, except that 
partial solution costs must be specified for each tuple.

\subsection{Cyclic Ordering}

Up until now, we have always used a graph representation of the problem 
instance. Some problems, however, can be represented more naturally by a 
hypergraph. An example is the \prob{Cyclic Ordering} problem, 
which takes 
as instance 
a set $V$, described by the predicate \verb!vertex/1!, 
and a set of triples described by \verb!order(A,B,C)!, where $A, B, C$ are 
declared vertices.  $f : V \to \{1,\dots,\lvert V \rvert\}$ is called an 
ordering of the vertices. A triple \verb!order(A,B,C)!  is said to be satisfied 
by an ordering $f$ iff $f(A) < f(B) < f(C)$ or $f(B) < f(C) < f(A)$ or $f(C) < 
f(A) < f(B)$.  The objective is to bring the vertices into an ordering such that 
all triples are satisfied.

The following monolithic encoding achieves this:
\begin{lstlisting}
1 { map(V,1..N) } 1 :- vertex(V), N = #count { vertex(_) }.
:- map(V1,K), map(V2,K), V1 < V2.
lt(V1,V2) :- map(V1,K1), map(V2,K2), K1 < K2.
sat(A,B,C) :- order(A,B,C), lt(A,B), lt(B,C).
sat(A,B,C) :- order(A,B,C), lt(B,C), lt(C,A).
sat(A,B,C) :- order(A,B,C), lt(C,A), lt(A,B).
:- order(A,B,C), not sat(A,B,C).
\end{lstlisting}
One can naturally represent the problem by a hypergraph where one considers the 
elements of $V$ as vertices and the triples as hyperedges. This ensures that the 
vertices in each triple must appear together in at least one bag and each triple 
can therefore be checked in an exchange node.

We use this opportunity to introduce a distinction between two ways of how to 
write exchange encodings: We can either do this, as we did in all the cases 
before, in a ``bottom-up'' way, i.e., by selecting preceding child tuples and 
then constructing the new tuples originating from them. When dealing with just a 
decision problem, another possibility is to proceed ``top-down'', i.e., by 
guessing first an assignment of values to current vertices and then checking if 
some child tuple exists that is a valid predecessor of the guess. This has the 
advantage that we avoid a guess over a potentially huge number of child tuples 
when probably many of the resulting tuples would coincide.

The following exchange encoding for \prob{Cyclic Ordering} proceeds 
``top-down'':
\begin{lstlisting}
1 { map(V,1..N) } 1 :- current(V), N = #count { current(_) }.
:- map(V1,K), map(V2,K), V1 < V2.
lt(V1,V2) :- map(V1,K1), map(V2,K2), K1 < K2.
sat(A,B,C) :- order(A,B,C), lt(A,B), lt(B,C).
sat(A,B,C) :- order(A,B,C), lt(B,C), lt(C,A).
sat(A,B,C) :- order(A,B,C), lt(C,A), lt(A,B).
:- order(A,B,C), current(A;B;C), not sat(A,B,C).
gtChild(I,V1,V2) :- mapped(I,V1,K1), mapped(I,V2,K2), current(V1;V2),    K1 > K2.
noMatch(I) :- lt(V1,V2), gtChild(I,V1,V2).
match :- childTuple(I), not noMatch(I).
:- not match.
\end{lstlisting}
The first seven lines are very similar to the monolithic encoding.  The 
remainder of the program makes sure that there is some valid predecessor among 
the child tuples.  Note that such an approach can only be employed for decision 
problems because counting or enumerating solutions require pointers to child 
tuples.
In fact, this problem differs from the previously discussed insofar as for those 
the set of values that can be mapped to a vertex was fixed (e.g., three colors 
or a truth value), whereas here it would be required to assign each vertex a 
number from $1$ up to the total number of vertices (just as the monolithic 
program does), if one were to really construct a solution to the problem
with this approach.  This 
would violate the principles of dynamic programming because subproblems would 
depend on the whole problem. However, note that in line~1, only a \emph{local} 
ordering, i.e., an ordering of the current bag elements, is guessed, so the 
value mapped to a vertex is \emph{not} its position in an ordering of all the 
vertices.  

\subsection{Further Examples and Overview}
\label{sec:further-examples}

In this section, we have illustrated the functioning of D-FLAT via several 
examples in the course of which we have also introduced step-by-step the special 
predicates D-FLAT provides or recognizes. 
D-FLAT 
solutions for 
further problems 
are provided on the system website at 
\url{http://www.dbai.tuwien.ac.at/research/project/dynasp/dflat/}.
Let it be noted here that it is also possible to handle problems higher on the 
polynomial hierarchy than \NP. For instance, one can solve (propositional) ASP itself with 
D-FLAT, for instance by implementing the algorithm presented in \cite{JaklPW09}.
For purposes like this, D-FLAT allows not only for assignments to all 
vertices but more generally for a hierarchy of assignments. For example, each 
top-level assignment (as we have been dealing with before) might have subsidiary 
assignments that are used to determine if the respective top-level assignment is 
valid.  A detailed account of this is beyond the scope of this paper. We refer 
the reader to the system website for examples of this.



Let us briefly summarize the main features of D-FLAT.
In many cases,
the user has to implement only the program for exchange nodes, while for 
join nodes D-FLAT offers a default implementation. For certain problems, this 
default has to be overridden. In their programs, users can exploit the full 
language accepted by Gringo.  The only restriction is that some predicates are 
reserved 
by D-FLAT.
Table~\ref{tab:special_predicates} gives a summary of these predicates and also 
points out under which circumstances (depending on the program or problem type) 
which predicates are used.
%
Predicates used in the facts that constitute the problem instance 
are supplied verbatim
to the exchange and join programs. 

\section{The D-FLAT System}

\subsection{System Description}
\label{sec:systemdescription}

\begin{figure}
\caption{A flowchart illustrating the (simplified) interplay of D-FLAT's 
components}
\label{fig:flowchart}
\input{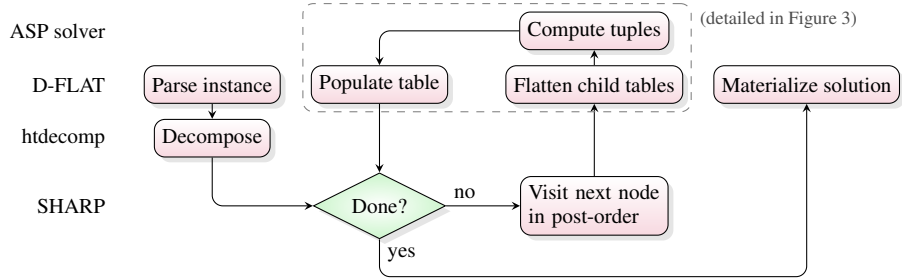}
\end{figure}

D-FLAT is written in C++ and can be compiled for many platforms. The 
main libraries used are: 
(1)
the SHARP framework\footnote{See \url{http://www.dbai.tuwien.ac.at/research/project/sharp/}.}
which takes care of constructing a tree decomposition of the input graph and 
then semi-normalizing it; as well, SHARP provides the skeleton for D-FLAT's data management;
(2)
SHARP itself uses the htdecomp library\footnote{See \url{%
http://www.dbai.tuwien.ac.at/proj/hypertree/downloads.html}.} 
which implements several heuristics for (hyper)tree decompositions, see also \cite{Dermaku08};
(3)
the 
Gringo/Clasp\footnote{See \url{http://potassco.sourceforge.net/}.}
family of ASP tools  (see also \cite{GebserKKOSS11} for a recent survey)
is finally used for grounding and solving the user-specified ASP programs. 
Figure~\ref{fig:flowchart} depicts the control flow between these components in 
a simplified way:
D-FLAT initially parses the instance and constructs a hypergraph representation, 
which is then used by htdecomp to build a hypertree decomposition. SHARP is now 
responsible for traversing the tree in post-order. For each node, it calls 
D-FLAT which flattens the child tables, i.e., converts them to a set of facts 
which is given to the ASP solver to compute the new tuples as answer sets.
These are collected by D-FLAT which populates the current node's table.  When 
all nodes have been processed like this, D-FLAT reconstructs the solutions from 
the tables.\footnote{%
The data flow during the procedure within a node (indicated by the dashed box) 
is illustrated in Figure~\ref{fig:dataflow}. Note that D-FLAT also features a 
default join implementation which does not use ASP and is not depicted.}

Since we leave the decomposition part as well as the solving of the ASP programs 
to external libraries, D-FLAT immediately takes advantage of improvements in 
these libraries. It is thus also possible to switch to another ASP solver 
altogether without changing \mbox{D-FLAT} internals except, of course, the parts calling 
the solver. Likewise, our approach allows to replace the tree decomposition  
part with weaker (but more efficient) concepts like graph cuts, etc.

Since the user provides ASP programs for the exchange and, optionally, join 
nodes at runtime, D-FLAT is independent of the particular problem being solved.  
This means that there is no need to recompile in order to change the dynamic 
programming algorithm. It can be used out of the box as a rapid prototyping tool 
for a wide range of problems. 

\begin{table}
\caption{D-FLAT's most important command-line options}
\label{tab:command-line-options}
\begin{tabular}{p{4cm}p{8cm}}
\hline
\hline
Argument & Meaning\\
\hline

\verb!-e edge_predicate! \newline \hspace*{1em} \footnotesize (mandatory at 
least once) &
\verb!edge_predicate! is used in the problem instance as the predicate declaring 
(hyper)edges. Arguments of this predicate implicitly declare the vertices of the 
input graph.\\
\noalign{\vspace{1ex}}

\verb!-x exchange_program! \newline \hspace*{1em} \footnotesize (mandatory) &
\verb!exchange_program! is the file name of the ASP program processing exchange 
nodes.\\
\noalign{\vspace{1ex}}

\verb!-j join_program! \newline \hspace*{1em} \footnotesize (optional) &
\verb!join_program! is the file name of the ASP program processing join nodes.  
If omitted, the default implementation is used.\\
\noalign{\vspace{1ex}}

\verb!-p problem_type! \newline \hspace*{1em} \footnotesize (optional) &
\verb!problem_type! specifies what kind of solution the user is interested in.  
It can be either ``enumeration'' (default), ``counting'', ``decision'', 
``opt-enum'', ``opt-counting'' or ``opt-value''.\\


\hline
\hline
\end{tabular}
\end{table}

When executing the D-FLAT binary, the user can adjust its behavior using 
command-line options. The most important ones are briefly described in 
Table~\ref{tab:command-line-options}.

The problem instance, which must be a set of facts, is read from the standard 
input. The parser of D-FLAT recognizes the predicates that declare (hyper)edges, 
whose names are given as command-line options, and for each such fact introduces 
a (hyper)edge into the instance graph. The arguments of the (hyper)edge 
predicates implicitly declare the instance graph's set of vertices. Note that 
the tree decomposition algorithm used currently does not allow for isolated vertices, but 
this is no real limitation because usually solutions of a modified instance
without the isolated vertices can be trivially extended.

Executing the D-FLAT binary \verb!dflat! is illustrated by the following example 
call, presupposing a \prob{3-Col} instance
with the file name \verb!instance.lp! and an exchange program
with the file name \verb!exchange.lp! (cf.\ Section~\ref{sec:graph-coloring}),
instructing D-FLAT 
to
print the number of solutions and enumerate them:
\begin{verbatim}
dflat -e edge -x exchange.lp < instance.lp
\end{verbatim}

%

\subsection{Experiments}
\label{sec:experiments}

In this section, we briefly report on first experimental results for the 
discussed problems.  We compared D-FLAT encodings to monolithic ones using 
Gringo and Clasp. Each instance was limited to 15~minutes of CPU time and 6~GB 
of memory. Instances were constructed such that they have small treewidth by 
starting with a tree-like structure and then introducing random elements until 
the desired fixed decomposition width is reached.


Traditional ASP solvers employ clever heuristics to quickly either find some 
model or detect unsatisfiability, thereby being able to solve the decision 
variant of problems particularly well. In contrast, 
the dynamic programming 
approach of 
D-FLAT currently always calculates the tuple tables in the same way, 
whatever the problem variant may be---it is only in the final materialization 
stage that solutions are assembled differently, depending on the problem type.

\prob{3-Col} and \prob{Sat} are prime examples of problems where traditional ASP 
solvers are very successful in solving the decision variant efficiently.  
However, when it no longer suffices to merely find \emph{some} model (e.g., when 
dealing with counting or enumeration problems), the decomposition exploited by 
D-FLAT pays off for small treewidths, especially when there is a great number of 
solutions. 
In our experiments, the monolithic encodings indeed 
soon hit the time limit. D-FLAT, on the other hand, even sooner ran out of 
memory for enumeration due to its materializing all solutions in memory at the 
same time, and because the number of solutions increased rapidly with larger 
instances.  This is a motivation to improve the materialization procedure in the 
future with incremental solving techniques.

\begin{figure}
\caption{Comparison of D-FLAT to monolithic encodings}
\subfloat[The decision and counting variant of \prob{Sat} on instances of 
treewidth 8]{%
\label{fig:sat-plot}%
\includegraphics[scale=0.4]{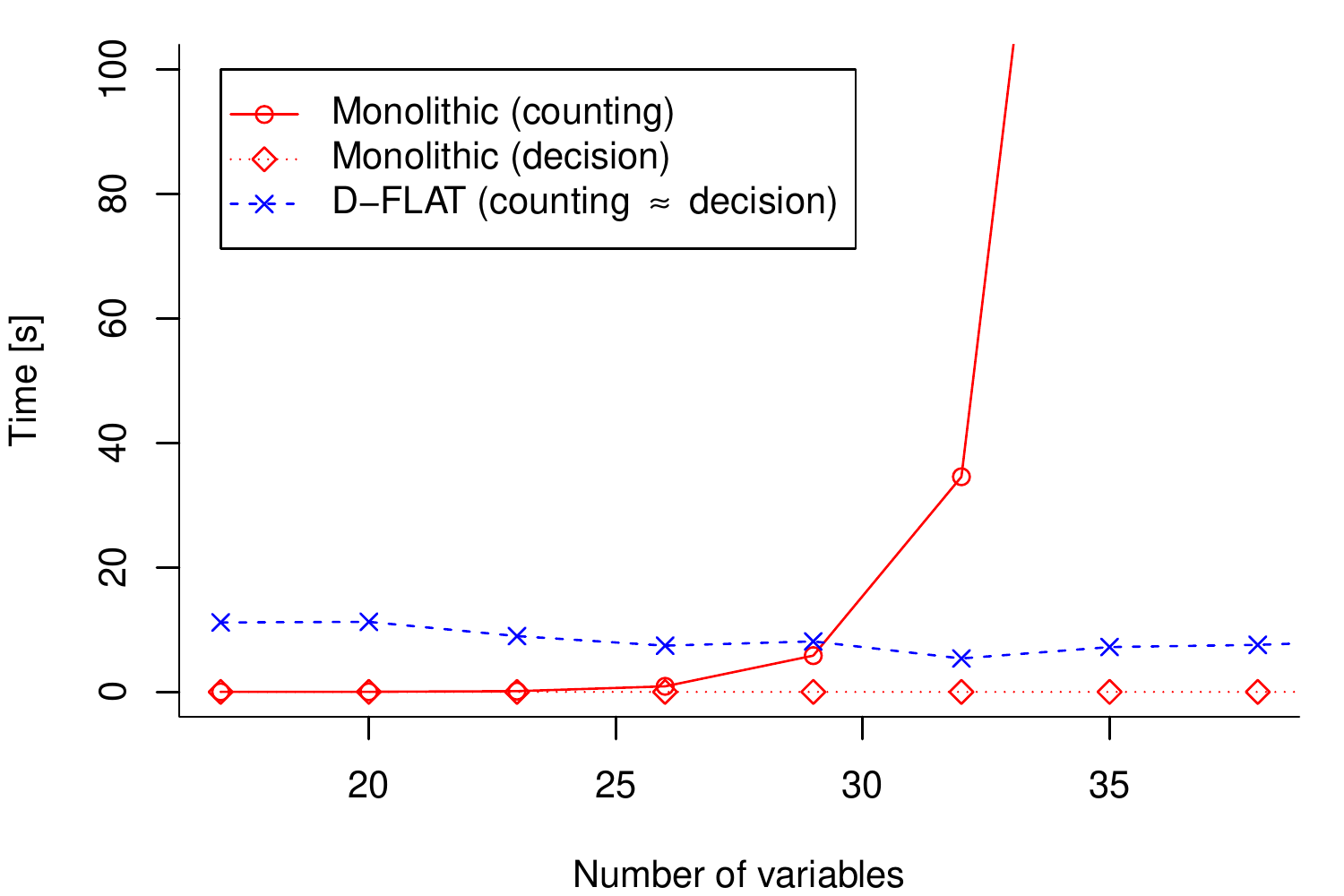}%
}%
\quad
\subfloat[Determining the optimum cost for \prob{Minimum Vertex Cover} on 
instances of treewidth 12]{%
\label{fig:vertex-cover-plot}%
\includegraphics[scale=0.4]{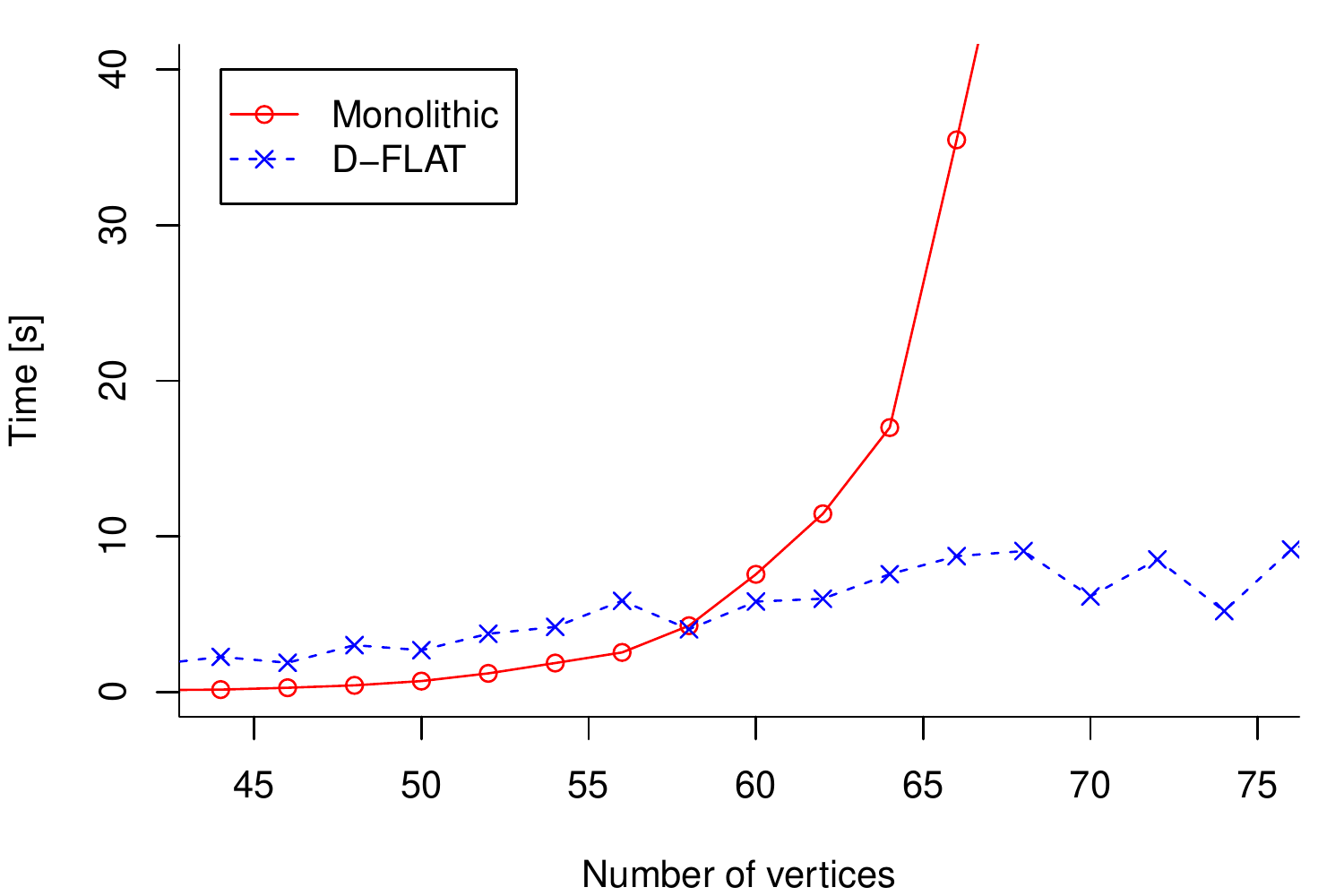}%
}
\end{figure}

Where D-FLAT excelled was the counting variant of these problems. It could solve 
each instance in a matter of seconds, while the monolithic program again ran out 
of time soon. Figure~\ref{fig:sat-plot} illustrates the strengths and weaknesses 
of D-FLAT for the counting and decision variants of \prob{Sat}: Although the 
monolithic program almost instantaneously solved the decision problem of all of 
the test instances, its running time on the counting problem soon exploded  
(standard ASP-solvers do not provide a dedicated functionality for counting and thus
have to implicitly enumerate all answer sets)
whereas D-FLAT remained almost unaffected on average.

Although most of the time traditional ASP-solvers 
perform 
very efficiently on decision problems, for some problems they have more difficulties, in particular 
when the grounding becomes 
huge.  Our investigations show that 
for 
the \prob{Cyclic Ordering} problem, D-FLAT often outperforms the monolithic 
program, but it could also be observed that D-FLAT's running time is heavily 
dependent on the constructed tree decomposition. For this reason, we averaged
over the performance on multiple tree decompositions for each instance size.

The \prob{Minimum Vertex Cover} problem proved to be a strong suit of D-FLAT 
(cf.\ Figure~\ref{fig:vertex-cover-plot}).  In optimization problems in general, 
stopping after the first solution has been found is not an option for 
traditional solvers, since yet undiscovered solutions might have lower cost.  
Another advantage of D-FLAT is that traditional solvers (at least in the case of 
Clasp) require two runs for counting or enumerating all optimal solutions: The 
first run only serves to determine the optimum cost, while the second starts 
from scratch and outputs all models having that cost.  D-FLAT, in contrast, only 
requires one run at the end of which it immediately has all the information 
needed to determine the optimum cost. 

As a concluding remark, recall that D-FLAT's main purpose is to provide a means 
to specify dynamic programming algorithms declaratively and not to compete with 
traditional ASP solvers, which is why we refrain from extensive benchmarks in 
this paper.  
%
Nonetheless,
it can be concluded that
D-FLAT is particularly successful for optimization and counting 
problems (provided the treewidth is small), especially when the number of 
solutions or the size of the monolithic grounding explodes. 

\section{Conclusion}
\label{sec:conclusion}

\paragraph*{Summary.}
We have introduced D-FLAT, a novel system which allows to specify 
dynamic programming algorithms over (hyper)tree decompositions by means of 
Answer-Set Programming.
To this end, D-FLAT employs an ASP system as an underlying inference
engine to compute the local solutions of the specified algorithm.
We have provided 
case studies illustrating how
the rich syntax of ASP allows for succinct and easy-to-read specifications
of such algorithms. 
The system%
---together with the example programs given in Section~\ref{sec:examples} 
and the benchmark instances used in Section~\ref{sec:experiments}---%
is free software and 
available at
\begin{quote}
\url{http://www.dbai.tuwien.ac.at/research/project/dynasp/dflat/}.
\end{quote}
%

\paragraph*{Related Work.}
Several forms of support of 
dynamic programming  exist in the world of PROLOG, where tabling 
of intermediate results is a natural concept. We mention here the 
systems TLP \cite{GuoG08}  and  B-Prolog \cite{DBLP:journals/corr/abs-1103-0812}.
In the world of datalog, we refer to
the Dyna system \cite{dyna} which provides a wide range of features 
for memoization. On the other side of the spectrum, 
the work of \cite{GottlobPW10}
shows that the simple formalism 
of monadic datalog can be placed instead of monadic second-order (MSO) logic
in meta-theorems (about fixed-parameter tractability in terms of 
treewidth) \`a la Courcelle.
The latter approach is thus more of theoretical interest.
%
The main difference between D-FLAT and the mentioned systems 
is that the user can make use of the full language of ASP and, in particular,
employ the Guess \& Check methodology to subproblems.
However, it has to be mentioned that our system so far supports only dynamic 
programming in connection with (hyper)tree decompositions.


\paragraph{Future Work.}
As a next step, we have to compare D-FLAT in more detail to the aforementioned 
approaches in terms of both modeling capacities and
performance issues. As well, we plan to provide different decomposition options
within D-FLAT. In particular, we anticipate that even a rather simple (but efficient)
decomposition of the graph, say a simple split into two parts, might in practice lead to 
performance gains over monolithic encodings. 
Another line of optimization concerns 
lazy evaluation  strategies, for 
which incremental ASP techniques \cite{GebserSS11} have been developed.
Finally, the relation of our approach to 
reactive ASP \cite{GebserGKS11} might provide interesting new research directions.

\paragraph*{Acknowledgments} Supported by Vienna University of Technology
special fund ``Innovative Projekte 9006.09/008''.

\bibliographystyle{acmtrans}
\bibliography{bibliography}

\end{document}